\documentclass[runningheads]{llncs}
\usepackage[T1]{fontenc}
\usepackage{graphicx}
\usepackage{amsmath}
\usepackage{amssymb}
\usepackage{booktabs}
\usepackage{hyperref}
\usepackage{cite}
\usepackage{makecell}
\usepackage{color}
\usepackage{algorithm}
\usepackage{algorithmic}
\usepackage{orcidlink}

\begin{document}

\title{Hybrid Quantum-MambaVision: A Quantum-enhanced State Space Model for Calibrated Mixed-type Wafer Defect Detection}
\titlerunning{Hybrid Quantum-MambaVision}

\author{
Satwik Sai Prakash Sahoo\inst{1}\orcidlink{0009-0002-1943-9323} \thanks{Code available at \url{https://github.com/satwiksps/wafer-mamba}}  \and 
Jyoti Prakash Sahoo\inst{2}\orcidlink{0000-0002-6273-6174} \and
Ting Wang\inst{3}\orcidlink{0000-0002-7223-8849} \and
Subrota Kumar Mondal \inst{4}\orcidlink{0000-0002-0008-7797}
}
\authorrunning{Satwik Sahoo et al.}

\institute{Odisha University of Technology and Research, Bhubaneswar, India \\
\email{sahoospsatwik@gmail.com} \and
National Taiwan University of Science and Technology, Taipei, Taiwan \\ 
\and
East China Normal University, Shanghai, China \\
\and Macau University of Science and Technology, Taipa, Macau
}

\maketitle

\begin{abstract}
Extracting actionable knowledge from industrial visual data is fundamentally bottlenecked by extreme class imbalance and the prohibitive computational complexity of modern foundation models. In semiconductor manufacturing, identifying multi-label wafer defects is a complex spatial data mining task where overlapping patterns obscure critical root-cause signals. While Vision Transformers (ViTs) excel at global dependency extraction, their quadratic scaling renders them inefficient for high-throughput, real-time anomaly detection. To overcome these computational barriers, this paper introduces Hybrid Quantum-MambaVision, a highly efficient architecture tailored for spatial knowledge discovery. We integrate a linear-complexity State-Space Model (SSM) backbone with a Parameterized Quantum Context Adapter (QCA) and Low-Rank Adaptation (LoRA). The Mamba backbone efficiently captures long-range spatial dependencies, while the quantum adapter maps compressed latent features into a high-dimensional Hilbert space to disentangle complex, overlapping signatures. On the highly imbalanced MixedWM38 dataset, Hybrid Quantum-MambaVision achieves exceptional multi-label classification performance, significantly reducing the error rate on complex multi-defect topologies compared to classical baselines. Crucially, the quantum regularizer acts as a profound uncertainty calibrator, substantially reducing Maximum Calibration Error (MCE) and minimizing expected false-positive costs. This work establishes a scalable Quantum-Classical hybrid paradigm for efficient representation learning in industrial data mining.

\end{abstract}

\section{Introduction}
As the semiconductor industry transitions to sub-nanometer fabrication nodes, Automated Optical Inspection (AOI) systems generate an exponential volume of data. While Wafer Bin Maps (WBMs) are crucial for diagnosing root causes of process failures, extracting actionable insights from these Wafer Map Defect Patterns (WMDPs) poses a formidable spatial data mining challenge. Systems must accurately detect anomalies ranging from subtle, high-frequency localized defects to expansive, low-frequency global patterns. 

Current knowledge discovery pipelines face two critical algorithmic bottlenecks. First, real-world fabrication data exhibits severe class imbalance; catastrophic global defects occur orders of magnitude less frequently than standard localized errors. Second, classifiers must routinely disentangle overlapping, multi-label defect signatures without overfitting to the majority classes.

Although Vision Transformers (ViTs)~\cite{dbkwzudmhgo20} excel at global dependency extraction, their quadratic $O(N^2)$ complexity scales poorly with high-resolution inputs, rendering them computationally prohibitive for real-time inference on active fabrication lines. To enable scalable, high-throughput data mining, we propose \textit{Hybrid Quantum-MambaVision}. This architecture seamlessly integrates the linear computational efficiency of State-Space Models (SSMs)~\cite{gd24} with the powerful feature-disentangling capabilities of Quantum Machine Learning (QML). 

Our primary contributions are:
\begin{enumerate}
\item \textit{Efficient Foundation Architecture:} A highly scalable MambaVision framework for WBM data mining, leveraging Low-Rank Adaptation (LoRA) to fine-tune deep spatial representations with minimal parameter updates.
\item \textit{Quantum Context Adapter (QCA):} A 4-qubit variational quantum circuit that bottlenecks latent features, acting as a global regularizer to separate complex, mixed-label defects.
\item \textit{Algorithmic Trustworthiness:} A rigorous empirical risk analysis demonstrating that the quantum bottleneck intrinsically calibrates uncertainty, sharply reducing Maximum Calibration Error (MCE) and expected operational costs.
\end{enumerate}

\section{Related Works}

\subsection{Evolution of Vision Architectures for Data Mining}
Since the debut of AlexNet~\cite{ksh12}, Convolutional Neural Networks (CNNs) have dominated computer vision, with recent advances modernizing these architectures through Transformer-inspired principles. For instance, ConvNeXt~\cite{lmwfdx22} broadened the kernels of ResNet~\cite{hzrs16}, RegNetY~\cite{rkghd20} introduced systematic design spaces, and EfficientNetV2~\cite{tl21} leveraged neural architecture search to optimize efficiency. However, traditional CNNs inherently struggle to capture the global image context necessary for identifying subtle, distributed anomalies. Vision Transformers (ViTs)~\cite{dbkwzudmhgo20} revolutionized the field by leveraging self-attention to expand receptive fields, though they traditionally require massive datasets, a hurdle mitigated by DeiT's~\cite{tcdmsj21} distillation-based training. To manage the prohibitive quadratic computational complexity of pure self-attention, researchers developed hierarchical and spatially separable architectures, including Swin Transformer~\cite{llchwzlg21}, Twins~\cite{ctwzrwxs21}, and PVT~\cite{wxlfsllls21}. Seeking the optimal balance between accuracy and throughput, contemporary designs increasingly favor hybrid approaches; models such as CoAT~\cite{xxct21}, CrossViT~\cite{cfp21}, NextViT~\cite{lxllwxwzp22}, EfficientFormer~\cite{lywhetwr22}, and FasterViT~\cite{hatamizadeh2024fastervit} successfully fuse CNN-like localized processing with global Transformer mechanisms.

\subsection{Knowledge Discovery in Wafer Defect Datasets}
In the domain of semiconductor manufacturing, researchers have increasingly deployed Transformers and attention mechanisms to capture the intricate, multi-scale spatial dependencies of wafer map defects~\cite{ww22}. While these global contextual models outperform standard CNNs, they frequently falter when confronting extreme class imbalances and multi-label topologies. Recognizing that isolated, single-defect maps fail to represent real-world fabrication complexities, Wang et al.~\cite{wxyzl20-1} introduced the MixedWM38 dataset, establishing a benchmark for evaluating overlapping spatial patterns. Subsequent architectures, such as the encoder-decoder WaferSegClassNet (WSCN)~\cite{nmsmm22}, achieved strong classification accuracy on these mixed typologies. A critical gap remains in the literature: many automated inspection frameworks entirely ignore mixed-type anomalies~\cite{sal20}, while others optimize strictly for accuracy at the expense of computational efficiency~\cite{clhzhw23, m24}. Given the immense throughput demands of real-world wafer factories, solutions that struggle to disentangle overlapping defects~\cite{sk23, chwhm24} or suffer from slow inference speeds~\cite{kts24} remain fundamentally unsuited for active industrial deployment.

\subsection{State-Space Models and Quantum Acceleration}
To overcome the computational bottlenecks of self-attention, State-Space Models (SSMs) have emerged as highly efficient and flexible sequence learners~\cite{jdsla24}. Mamba~\cite{gd24}, in particular, captures long-range dependencies with strictly linear time complexity, enabling rapid inference under constrained hardware conditions~\cite{sdz24}. While vision-centric adaptations like VMamba~\cite{ltzyxwyl24}, EfficientVMamba~\cite{phx24}, and SiMBA~\cite{pa24} successfully translated SSMs to spatial tasks, they often introduce computational overhead or constrain receptive fields. MambaVision resolves these limitations by strategically applying CNNs at high resolutions and SSMs at deeper semantic bottlenecks, maximizing both accuracy and throughput. As Moore’s law decelerates, handling increasingly massive, high-dimensional datasets demands unconventional computing paradigms. Quantum computing offers an alternative, leveraging superposition and high-dimensional tensor product spaces to accelerate complex optimization tasks, as demonstrated in quadratic unconstrained binary optimization (QUBO)~\cite{dpsp19}, protein folding~\cite{dosw08}, and graph clustering~\cite{s07}. Following Google's demonstration of quantum supremacy~\cite{aabbbbbbbbo19}, formulating machine learning bottlenecks to exploit these quantum computational advantages presents a profound frontier for scaling efficient foundation models.

\section{Dataset and Problem Formulation}
Mining overlapping spatial patterns is notoriously difficult. To rigorously evaluate our proposed architecture, we utilize the MixedWM38 dataset, an industrial benchmark containing both single and multi-label defect patterns.

\begin{figure}[!h]
\centering
\includegraphics[width=0.8\textwidth]{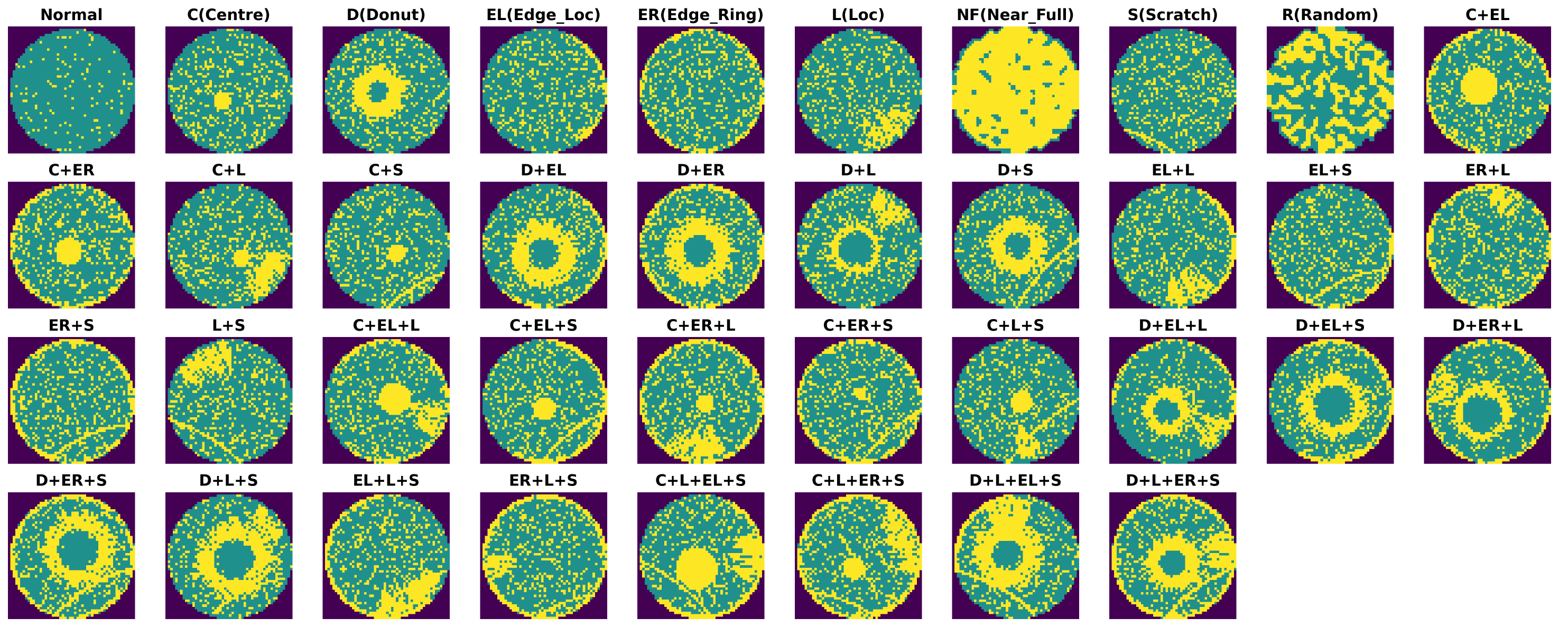}
\caption{Visual representation of distinct defect classes and mixed-label spatial topologies.}
\label{fig:wafer_defects}
\end{figure}

 Let $x_i \in \mathbb{R}^{H \times W \times 3}$ represent a WBM image and $y_i \in \{0, 1\}^C$ denote its corresponding multi-hot label vector across $C=8$ distinct defect classes. As shown in Fig.~\ref{fig:wafer_defects}, the spatial combinations of these anomalies create highly complex topologies, where dominant global defects often obscure subtle localized anomalies.

\begin{figure}[h]
\centering
\includegraphics[width=0.5\textwidth]{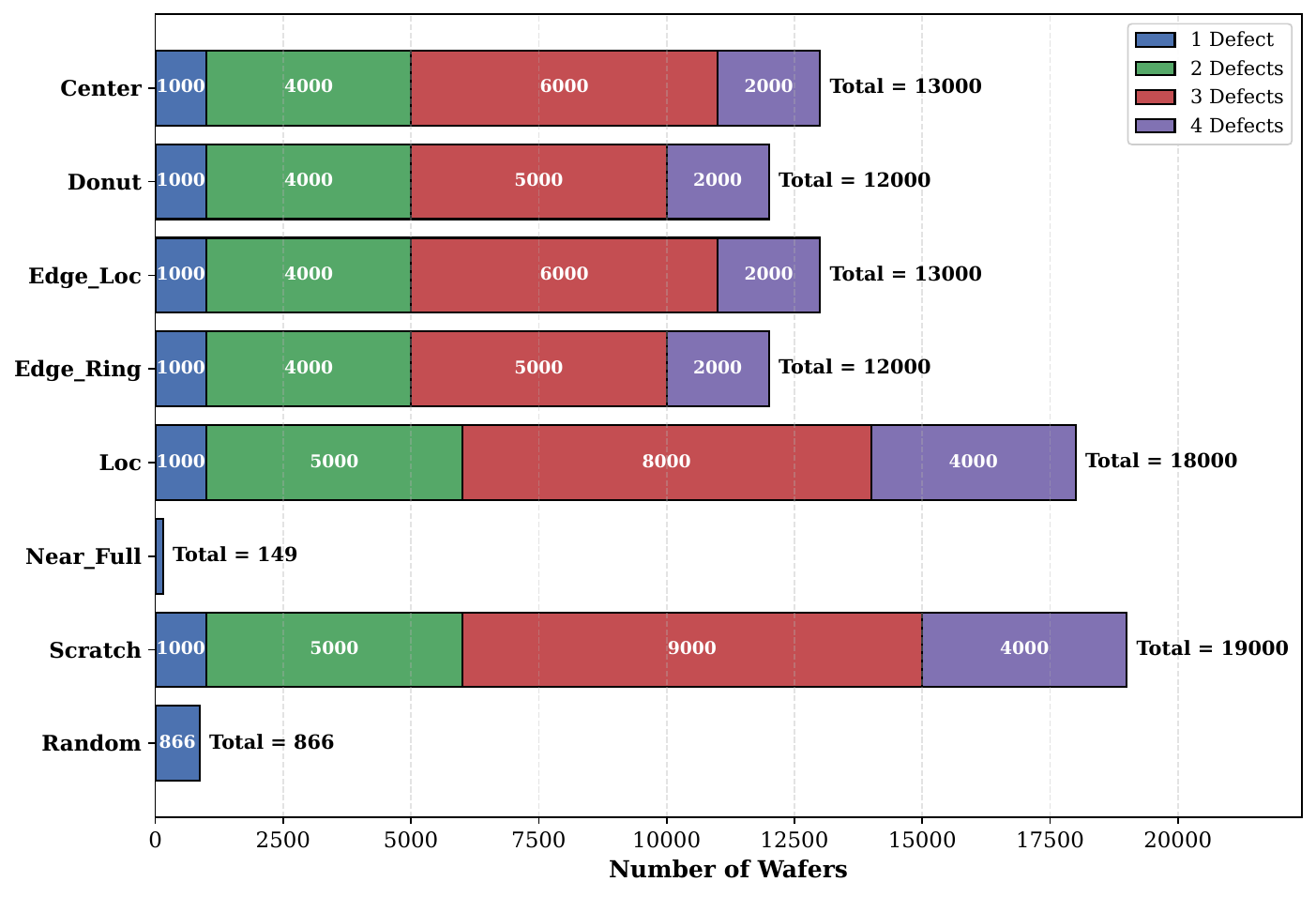}
\caption{Distribution of wafer defects, highlighting the extreme rarity of the `Near\_Full' class compared to standard localized defects.}
\label{fig:defect_dist}
\end{figure}

The primary challenge in this data mining task is extreme class imbalance. In active fabrication environments, defect occurrences follow a severe long-tail distribution. As illustrated in Fig.~\ref{fig:defect_dist}, standard defect patterns such as `Edge\_Ring' appear with high frequency, whereas critical macro-anomalies like `Near\_Full' consist of only 149 samples. An efficient foundation model must simultaneously extract these rare spatial frequencies without succumbing to majority-class bias.

\section{Methodology}
We propose a unified framework that synergizes the linear-time spatial modeling of Mamba with the expressivity of parameterized quantum circuits.

\begin{figure}[!ht]
\centering
\includegraphics[width=0.75\textwidth]{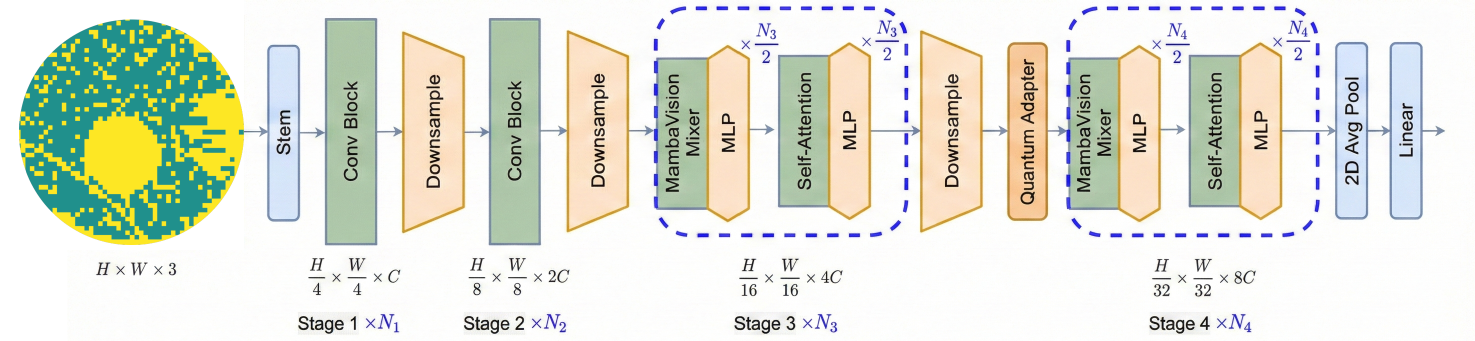}
\caption{The Hybrid Quantum-MambaVision Architecture. The network processes visual data through CNN stems and Mamba blocks, utilizing a Quantum Context Adapter to recalibrate features at the deepest semantic bottleneck.}
\label{fig:overall_arch}
\end{figure}
\begin{algorithm}[!ht]
\caption{Mamba Block Forward Pass}
\label{alg:mamba}
\begin{algorithmic}[1]
\REQUIRE Flattened input token sequence $X \in \mathbb{R}^{L \times D}$, State dimension $N$
\STATE $x, z \leftarrow \text{Linear}_{in}(X)$ \COMMENT{Input projection and gating branch}
\STATE $x' \leftarrow \text{Conv1D}(x)$ \COMMENT{Local spatial feature extraction}
\STATE $\Delta, B, C \leftarrow \text{Linear}_{proj}(x')$ \COMMENT{Data-dependent state parameters}
\STATE $A_d \leftarrow \exp(\Delta A)$ \COMMENT{Discretize state matrix}
\STATE $B_d \leftarrow (\Delta A)^{-1}(\exp(\Delta A) - I) \cdot \Delta B$ \COMMENT{Discretize input matrix}
\STATE $h_0 \leftarrow 0$
\FOR{$t = 1$ to $L$}
    \STATE $h_t \leftarrow A_d h_{t-1} + B_d x'_t$ \COMMENT{State-space recurrence loop}
    \STATE $y_t \leftarrow C h_t$ \COMMENT{Compute sequence output}
\ENDFOR
\STATE $y \leftarrow y \odot \text{SiLU}(z)$ \COMMENT{Apply gating mechanism}
\STATE $X_{out} \leftarrow \text{Linear}_{out}(y) + X$ \COMMENT{Output projection and residual connection}
\RETURN $X_{out}$
\end{algorithmic}
\end{algorithm}
\subsection{The Efficient MambaVision Backbone}
We anchor our framework on the MambaVision-T-1K architecture (Fig.~\ref{fig:overall_arch}). At its core, the continuous-time State-Space Model (SSM) transforms a 1-D input sequence into an output representation via a hidden latent state. To execute this system on digital hardware, the model discretizes the continuous state matrices using a Zero-Order Hold (ZOH) mapping driven by a timescale parameter, $\Delta$. Algorithm~\ref{alg:mamba} outlines the precise step-by-step discrete token processing. 

By structuring the visual sequence processing in this manner, the architecture achieves a global receptive field with strictly linear $O(N)$ computational complexity. This efficient scaling profile is crucial for mining macro-level defect topologies across high-resolution wafer maps, completely bypassing the prohibitive quadratic memory bottlenecks that limit standard Vision Transformers.

\subsection{Parameter-Efficient Fine-Tuning via LoRA}

\begin{figure}[!ht]
\centering
\includegraphics[width=0.5\textwidth]{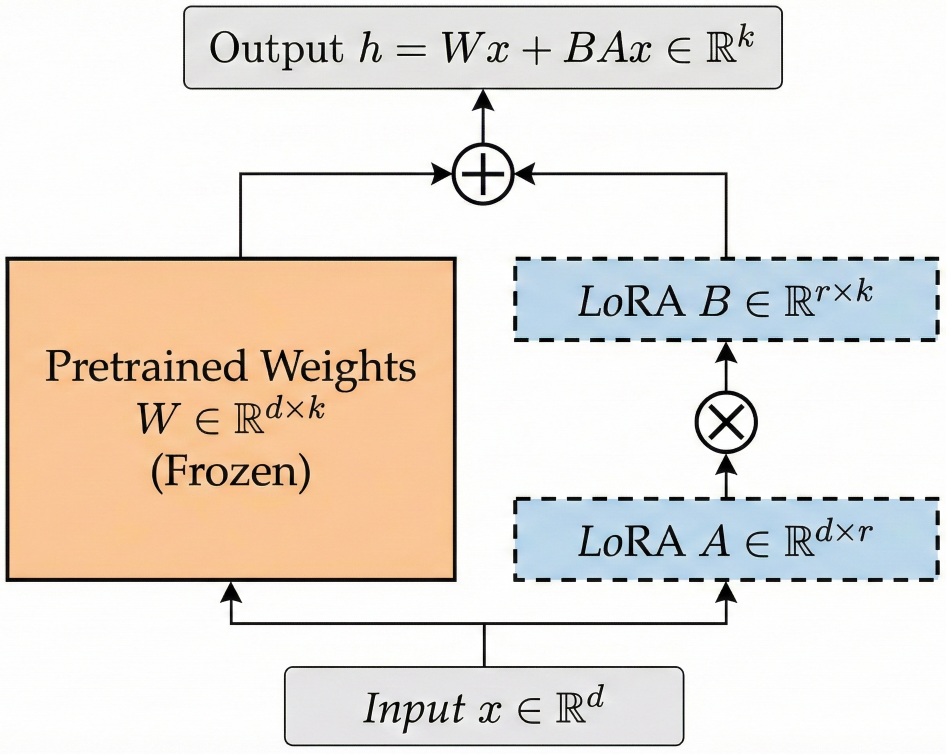}
\caption{Low-Rank Adaptation (LoRA) strategy applied to Mamba projections.}
\label{fig:lora}
\end{figure}

Fine-tuning massive foundation models on specialized industrial data frequently triggers catastrophic forgetting, degrading their generalized feature representations. To prevent this, we apply Low-Rank Adaptation (LoRA), operating on the hypothesis that the intrinsic dimensionality of weight updates required for wafer defect mapping is strictly low-rank. 

As illustrated in Fig.~\ref{fig:lora}, the pre-trained weight matrix $W_0 \in \mathbb{R}^{d \times k}$ remains frozen. The LoRA mechanism bypasses the frozen pre-trained weights, instead routing the input x through two low-rank matrices (A and B). This parallel pathway allows the model
to learn task-specific spatial dynamics for wafer defect mapping without triggering catastrophic
forgetting. During the forward pass, we instead inject a trainable low-rank decomposition $\Delta W = BA$, where $B \in \mathbb{R}^{d \times r}$, $A \in \mathbb{R}^{r \times k}$, and rank $r \ll \min(d, k)$:
\begin{equation}
h = W_0 x + \Delta W x = W_0 x + \frac{\alpha}{r} B A x
\end{equation}

By targeting the linear projections and fully connected layers with a rank of $r=64$ and a scaling factor of $\alpha=128$, the architecture robustly adapts its state-space dynamics to industrial WBMs while optimizing only a negligible fraction of the total parameters.

\subsection{Quantum Context Adapter (QCA)}
\begin{figure}[!h]
\centering
\includegraphics[width=0.5\textwidth]{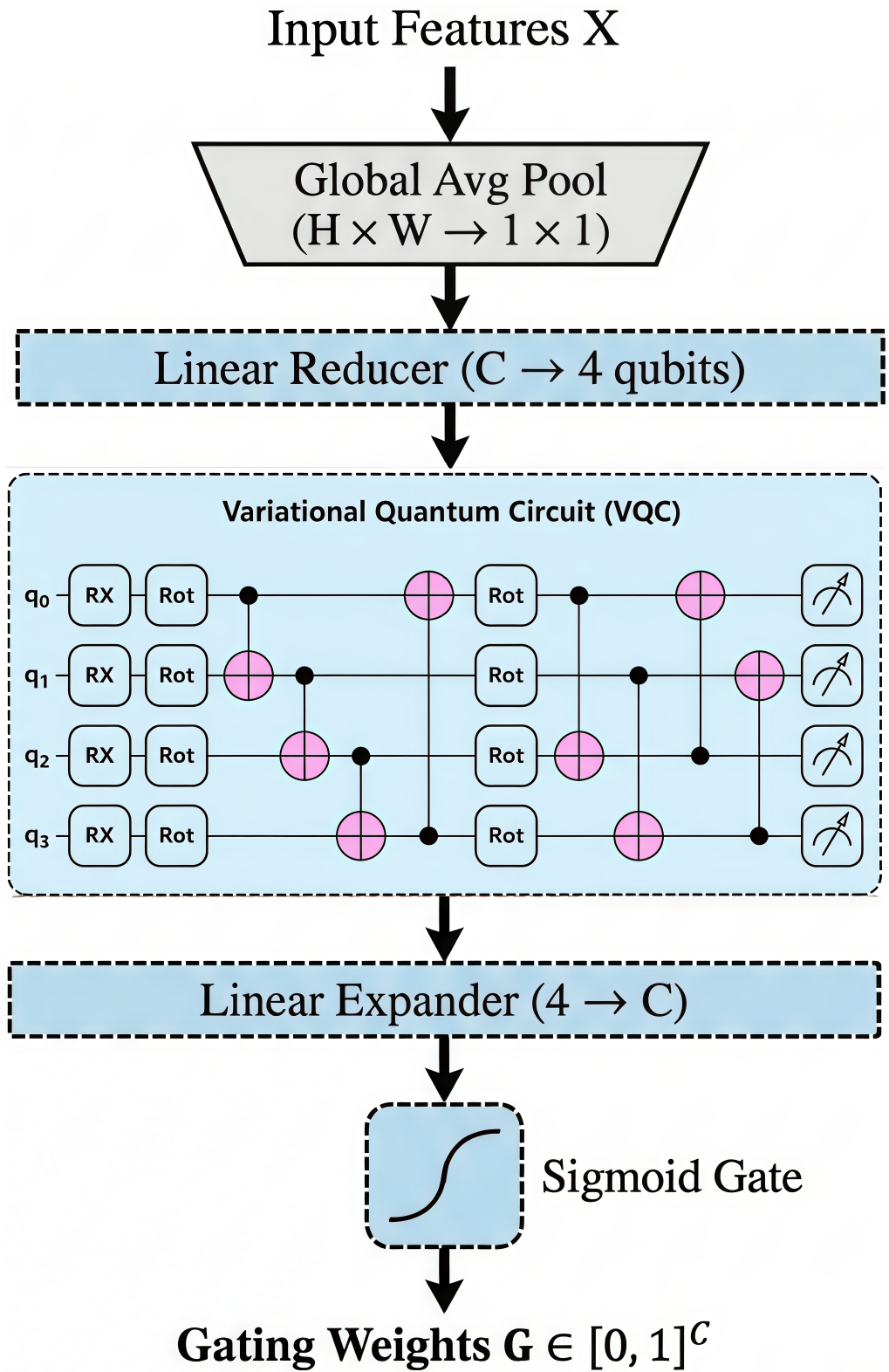}
\caption{The 4-qubit Quantum Context Adapter (QCA) architecture.}
\label{fig:qca}
\end{figure}
The core algorithmic novelty is the Quantum Context Adapter (Fig.~\ref{fig:qca}) inserted between Stage 3 and Stage 4. It acts as a strict informational bottleneck, mapping overlapping spatial dependencies into a high-dimensional quantum latent space. 

The specific step-by-step tensor operations and quantum measurements are defined in Algorithm~\ref{alg:qca}.

\begin{algorithm}[!ht]
\caption{Quantum Context Adapter (QCA) Forward Pass}
\label{alg:qca}
\begin{algorithmic}[1]
\REQUIRE Bottleneck features $X \in \mathbb{R}^{B \times C \times H \times W}$
\STATE $v \leftarrow \text{GAP}(X)$ \COMMENT{Global Average Pool to $\mathbb{R}^{B \times C}$}
\STATE $\phi \leftarrow \frac{\pi}{2} \tanh(W_{red} v)$ \COMMENT{Reduce to $n_q$ dimensions and scale}
\STATE $|\psi_0\rangle \leftarrow \bigotimes_{i=1}^{n_q} R_x(\phi_i)|0\rangle$ \COMMENT{Apply Angle Embedding}
\FOR{$l = 1$ to $L$}
    \STATE Apply $U(\theta_l)$ and CNOT gates \COMMENT{Strongly Entangling Layers}
\ENDFOR
\STATE $y_q \leftarrow \langle \psi_L | \sigma_z | \psi_L \rangle$ \COMMENT{Pauli-Z Measurement}
\STATE $G \leftarrow \sigma(W_{exp} y_q)$ \COMMENT{Classical expansion to $C$ channels}
\STATE $X_{out} \leftarrow X \odot (1 + \lambda \cdot G)$ \COMMENT{Residual spatial gating}
\RETURN $X_{out}$
\end{algorithmic}
\end{algorithm}

The Global Average Pooling (GAP) layer first compresses the spatial dimensions, and a reduction layer scales these classical values. We specifically select a 4-qubit architecture to optimally balance the dimensionality of the latent bottleneck against the computational overhead of quantum simulation. These reduced features are encoded into the quantum state using Angle Embedding ($R_x$ gates), which efficiently maps continuous classical variables into quantum amplitudes. The state then passes through Strongly Entangling Layers consisting of parameterized rotations and CNOT gates, chosen to maximize feature mixing and expressivity within the Hilbert space using minimal circuit depth. Finally, Pauli-Z measurements collapse the quantum state back into classical expectation values.

As depicted in Figure ~\ref{fig:qca}, the classical feature maps are first pooled and reduced before being
encoded into the quantum state. Following the parameterized operations, the Pauli-Z measurement
outputs are classically expanded and applied as a residual spatial gating mechanism, dynamically
recalibrating the original bottleneck features.

By constraining the global context into a 4-qubit representation, the Variational Quantum Circuit (VQC) maps non-linear correlations through unitary transforms. The learnable scaling parameter $\lambda$ initializes at $0$, ensuring the network maintains stable classical representations before gradually integrating the quantum context as an adaptive frequency filter.

\section{Experiments and Results}

\subsection{Setup and Training Dynamics}
We evaluated the Hybrid-Mamba (Hybrid Quantum-MambaVision) model against three established baselines: Classical MambaVision (strict ablation), ResNet-50 (local CNN benchmark), and ViT-Small (global Transformer benchmark). To counter the extreme class imbalance inherent in the WBM data, training utilized Focal Loss, structurally defined as:
\begin{equation}
L_{FL} = - \sum_{c=1}^C \alpha_c (1 - \hat{p}_c)^\gamma \log(\hat{p}_c)
\end{equation}
where $\gamma = 2$ aggressively penalizes prediction errors on rare multi-label classes. 

\begin{figure}[!ht]
\centering
\includegraphics[width=0.4\textwidth]{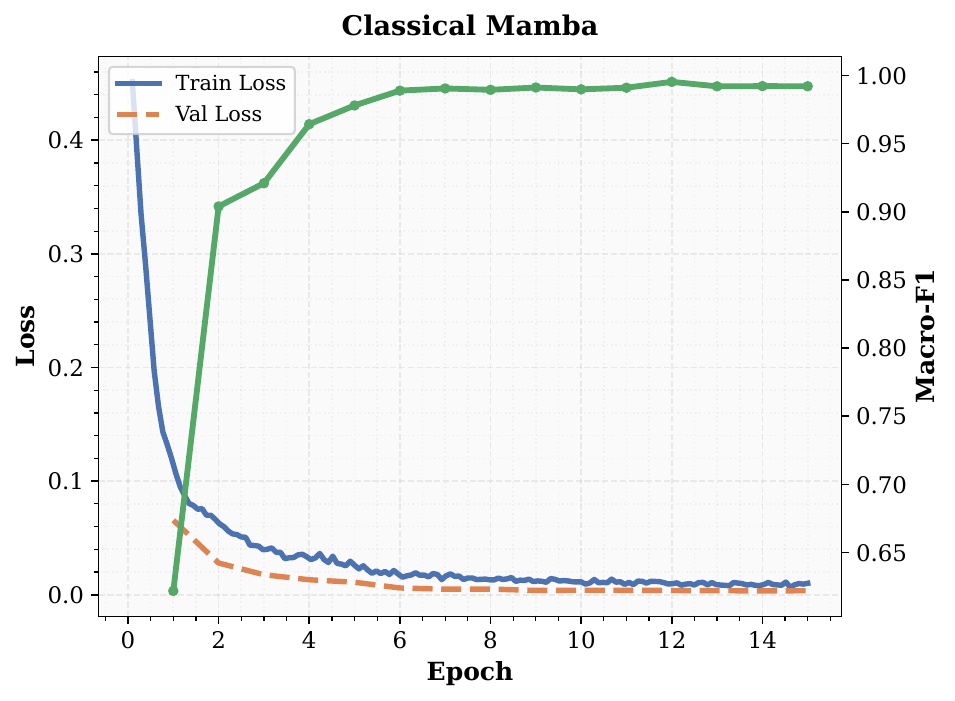}\hfill
\includegraphics[width=0.4\textwidth]{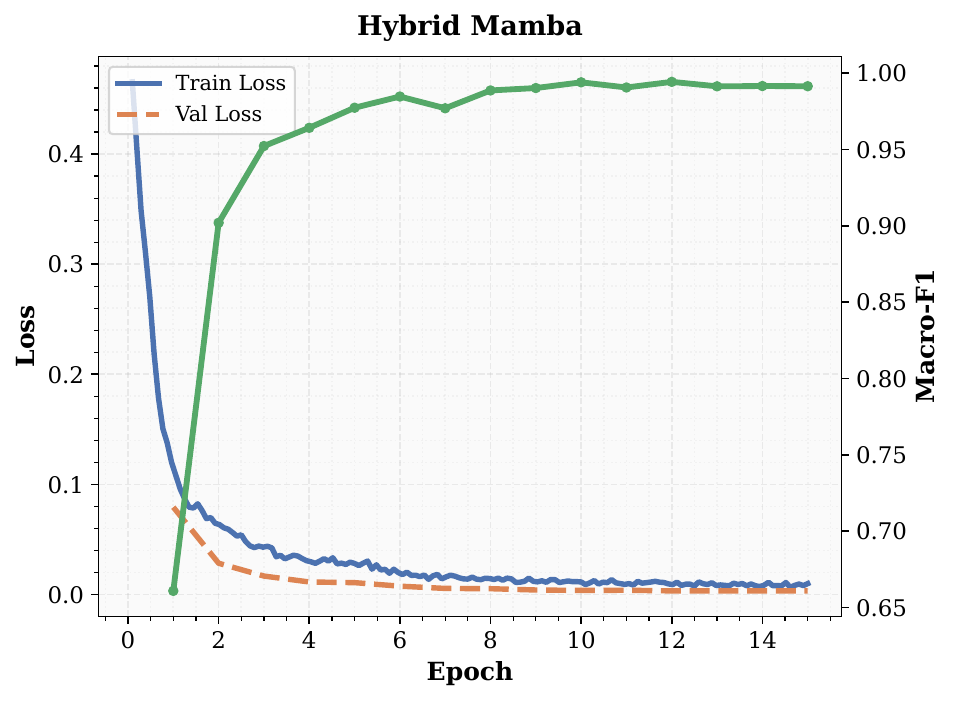}

\vspace{1em}

\includegraphics[width=0.4\textwidth]{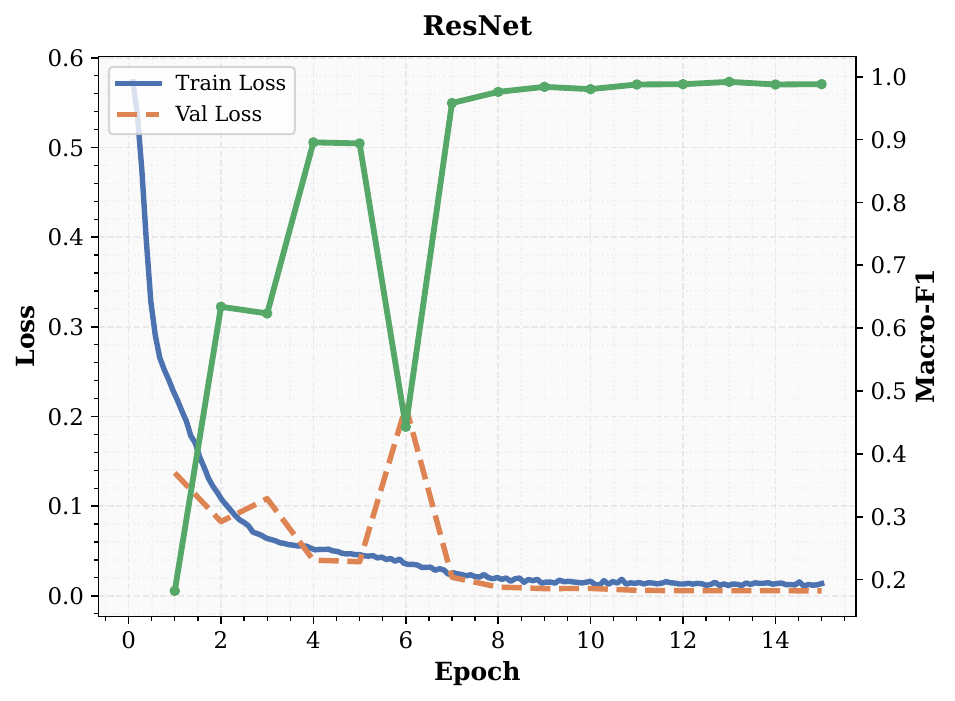}\hfill
\includegraphics[width=0.4\textwidth]{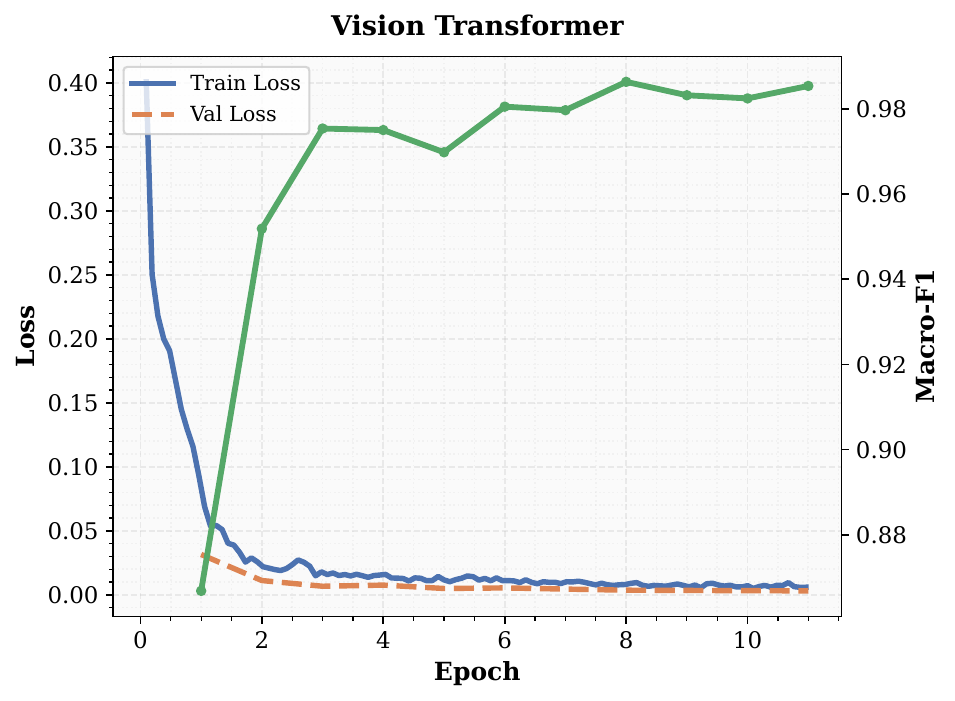}
\caption{Comparative training dynamics across the evaluated architectures. The Hybrid-Mamba model demonstrates rapid stabilization and smooth convergence without the early-epoch volatility observed in the classical ResNet and ViT baselines.}
\label{fig:training_dynamics}
\end{figure}

As shown in Fig.~\ref{fig:training_dynamics}, the Hybrid-Mamba model exhibits highly stable convergence. The inclusion of the VQC does not introduce vanishing gradients; instead, it rapidly smooths the validation loss curve.

\subsection{Multilabel Defect Classification Performance}
To rigorously evaluate the models on the MixedWM38 dataset, we analyzed both global multilabel metrics and per-class performance. 

\begin{table}[!ht]
\centering
\caption{Advanced Multilabel Quality Metrics across all architectures.}
\label{tab:multilabel}
\resizebox{\textwidth}{!}{
\begin{tabular}{lcccccc}
\toprule
\textbf{Model} & \textbf{mAP } & \textbf{Hamming Loss } & \textbf{Ranking Loss } & \textbf{Coverage Error } & \textbf{Kendall $\tau$ } \\
\midrule
Classical Mamba    & 0.99595 & 0.00437 & 0.00069 & 2.30786 & 0.99830 \\
\textbf{Hybrid Mamba} & \textbf{0.99727} & \textbf{0.00388} & \textbf{0.00056} & \textbf{2.30615} & 0.99841 \\
ResNet             & 0.99450 & 0.00723 & 0.00097 & 2.31115 & \textbf{0.99911} \\
Vision Transformer & 0.99549 & 0.00404 & \textbf{0.00056} & 2.30629 & 0.99816 \\
\bottomrule
\end{tabular}
}
\end{table}

As detailed in Table~\ref{tab:multilabel}, Hybrid-Mamba achieves the highest mean Average Precision (mAP) of 0.99727 while minimizing the Hamming Loss, indicating superior multi-label exact matches. This global superiority is driven by its exceptional performance on critically imbalanced minority classes.

\subsection{Resilience to Defect Complexity}
We evaluated the structural integrity of the models across subsets containing 1 to 4 simultaneous defects to test spatial entanglement resilience.

\begin{figure}[!ht]
\centering
\begin{minipage}{0.45\textwidth}
  \centering
  \includegraphics[width=\linewidth]{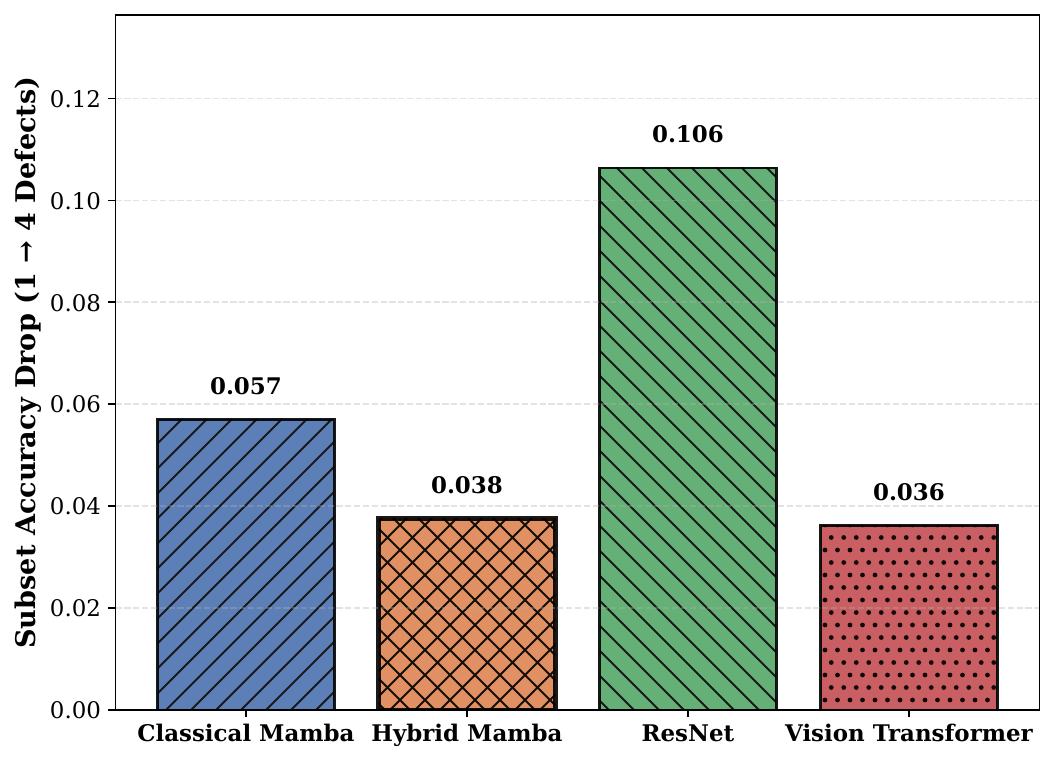}
  \caption{Accuracy degradation as concurrent defects increase from 1 to 4.}
  \label{fig:acc_drop}
\end{minipage}\hfill
\begin{minipage}{0.45\textwidth}
  \centering
  \includegraphics[width=\linewidth]{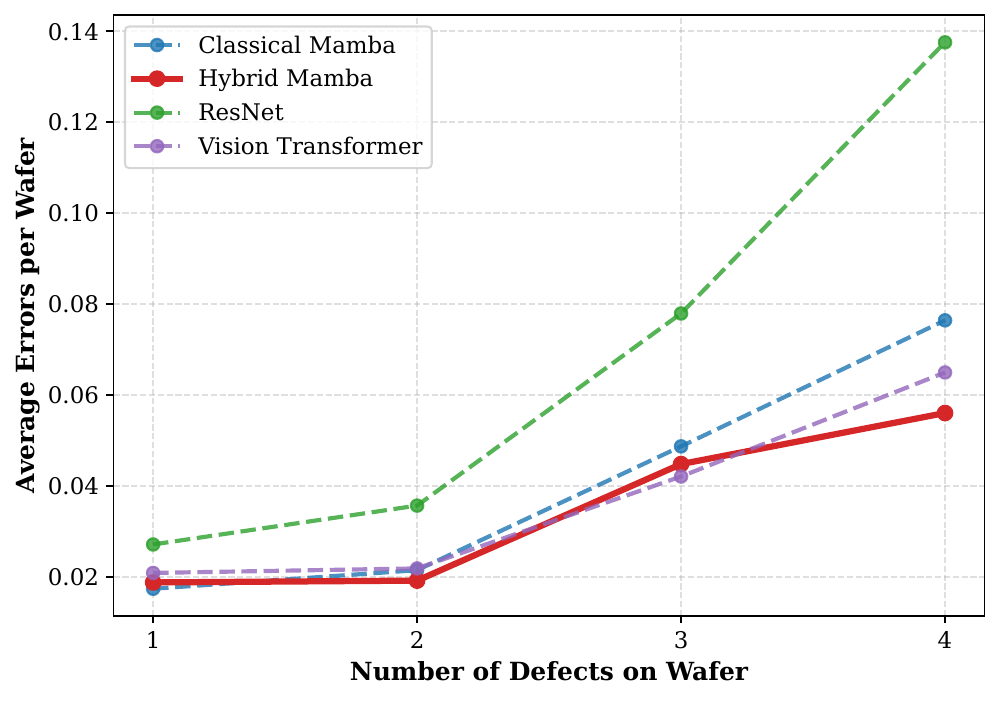}
  \caption{Error accumulation per wafer due to overlapping spatial frequencies.}
  \label{fig:err_accum}
\end{minipage}
\end{figure}

From Fig.~\ref{fig:acc_drop} we observe, ResNet-50 suffers a massive 0.106 drop in accuracy when transitioning to 4-defect wafers, as localized convolutions fail to parse intersecting boundaries. The Hybrid Mamba restricts this drop to just 0.038. Consequently, the error accumulation per wafer (Fig.~\ref{fig:err_accum}) is minimized, demonstrating the QCA's ability to successfully disentangle mixed signals in the latent space.

\subsection{Quantum Routing and Trustworthiness}
To understand the functional impact of the quantum layer on knowledge discovery, we analyzed the average quantum gate activation, defined as the magnitude of the scaling multiplier $\lambda G$ applied via the residual connection.

\begin{figure}[!h]
\centering
\includegraphics[width=0.8\textwidth]{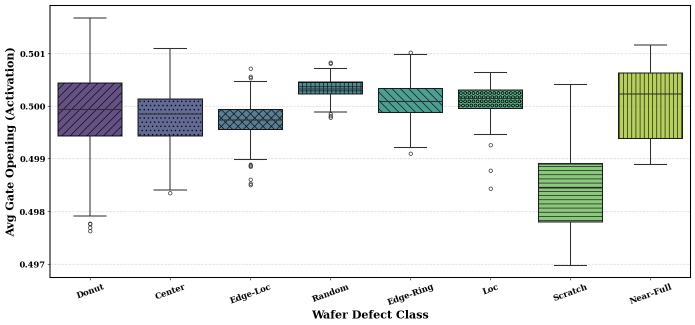}
\caption{Distribution of quantum gate activations across defect classes. }
\label{fig:gate_act}
\end{figure}

As illustrated in Fig.~\ref{fig:gate_act}, the classical optimizer actively utilizes the QCA rather than pruning it. The gate activation acts as a topology-aware router. For simple, unambiguous defects, the model relies heavily on the classical backbone. This quantum routing acts as a strict regularizer against algorithmic overconfidence. Classical deep learning models frequently suffer from blind overconfidence, necessitating an analysis of their calibration.

\begin{table}[!h]
\centering
\caption{Calibration and Confidence Quality Metrics.}
\label{tab:calibration}
\resizebox{\textwidth}{!}{
\begin{tabular}{lcccccc|c}
\toprule
\textbf{Model} & \textbf{ECE } & \textbf{MCE } & \textbf{Brier } & \textbf{NLL } & \textbf{Acc@Conf$\geq$0.9 } & \textbf{Entropy } & \textbf{Ambiguous \% } \\
\midrule
Classical Mamba    & \textbf{0.0398} & 0.6597 & 0.0064 & \textbf{0.0502} & 0.9709 & \textbf{0.15380} & 0.15288 \\
\textbf{Hybrid Mamba} & 0.0407 & \textbf{0.5553} & \textbf{0.0063} & 0.0503 & \textbf{0.9749} & 0.15499 & \textbf{0.12987} \\
ResNet             & 0.0554 & 0.5623 & 0.0118 & 0.0723 & 0.9478 & 0.19077 & 0.31562 \\
Vision Transformer & 0.0475 & 0.6204 & 0.0070 & 0.0578 & 0.9727 & 0.18035 & 0.12822 \\
\bottomrule
\end{tabular}
}
\end{table}

\begin{figure}[!h]
\centering
\begin{minipage}{0.32\textwidth}
  \centering
  \includegraphics[width=\linewidth]{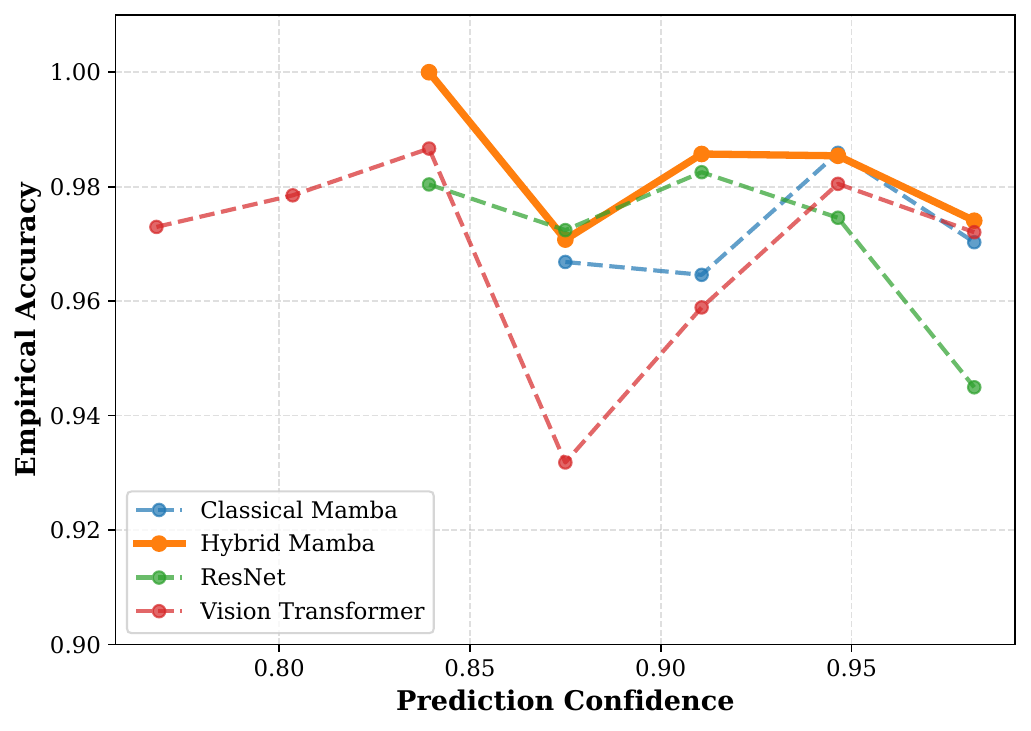}
  \caption{Empirical Accuracy vs. Prediction Confidence .}
  \label{fig:calibration}
\end{minipage}\hfill
\begin{minipage}{0.32\textwidth}
  \centering
  \includegraphics[width=\linewidth]{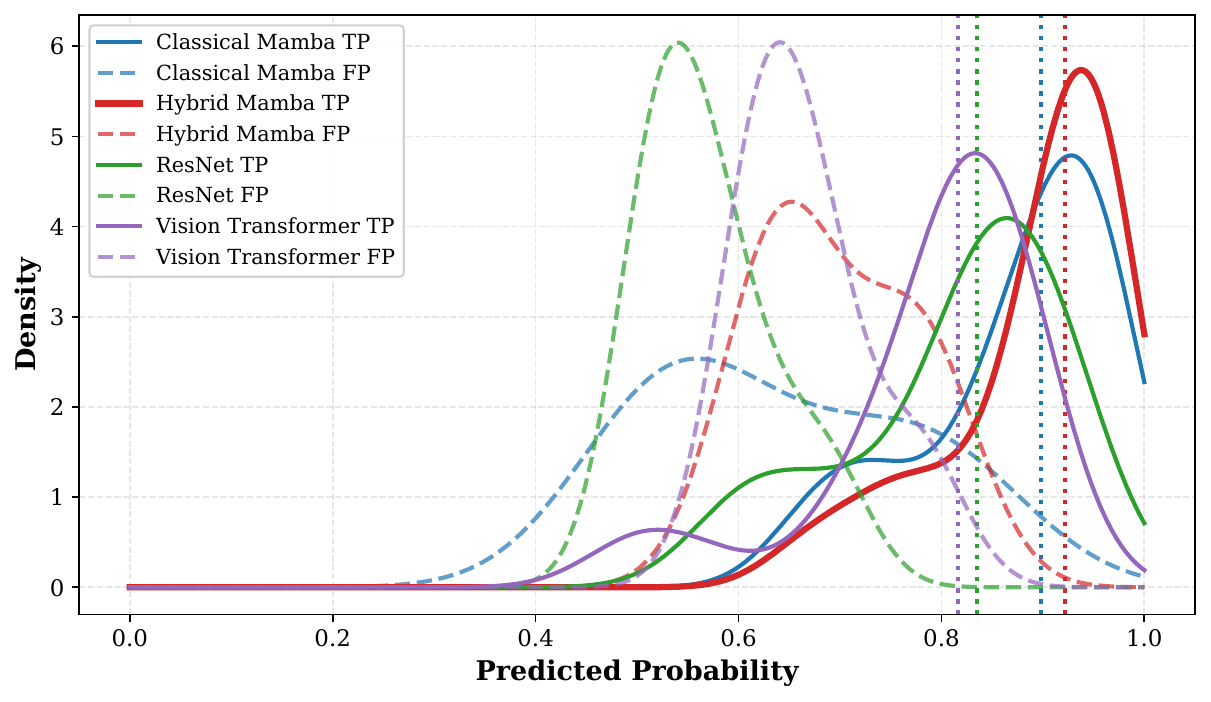}
  \caption{Confidence separation density for the highly imbalanced `Near\_Full' defect.}
  \label{fig:rare_sep}
\end{minipage}\hfill
\begin{minipage}{0.32\textwidth}
  \centering
  \includegraphics[width=\linewidth]{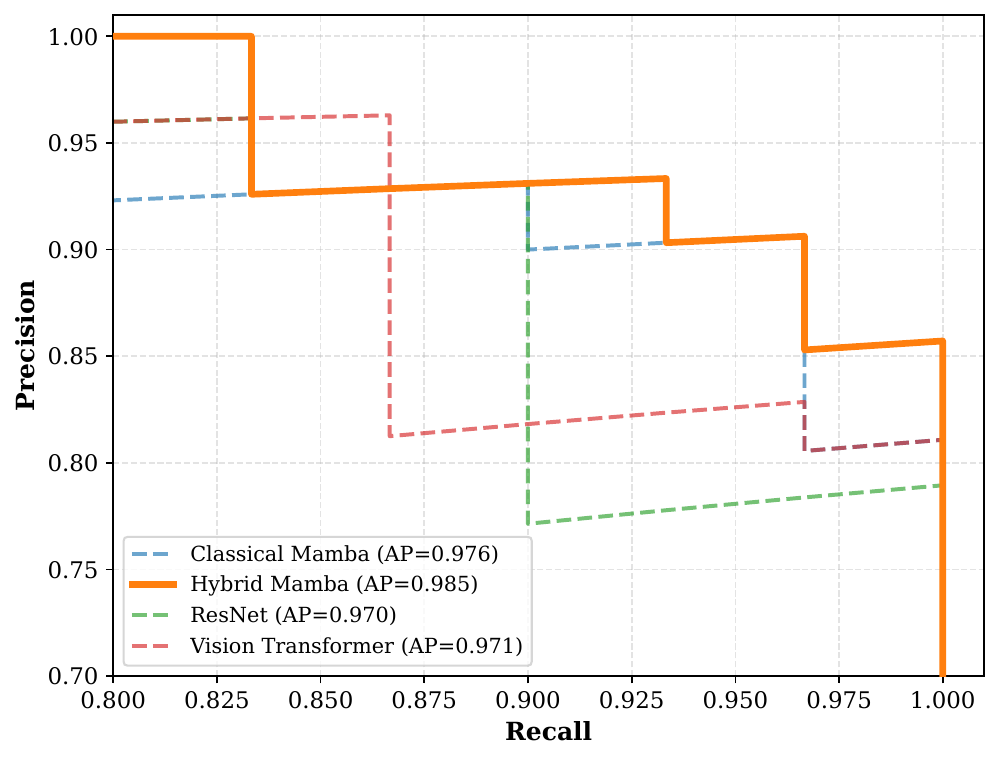}
  \caption{Precision-Recall curve for `Near\_Full', demonstrating an exceptional AP of 0.985.}
  \label{fig:pr_curve}
\end{minipage}
\end{figure}

As shown in Table~\ref{tab:calibration} and the reliability diagram (Fig.~\ref{fig:calibration}), the Classical Mamba architecture exhibits severe miscalibration (Maximum Calibration Error = 0.6597). By penalizing blind certainty on overlapping frequencies, the Hybrid Mamba drastically reduces the MCE to 0.5553 and decreases the percentage of ambiguous predictions to just 12.9\%. This improved trustworthiness is crucial for discovering rare classes, also demonstrated by the exceptional separation and Precision-Recall characteristics for the `Near\_Full' defect (Fig.~\ref{fig:rare_sep} and Fig.~\ref{fig:pr_curve}).

\subsection{Risk-Aware Deployment}
In real-world data mining applications, an efficient model must support selective prediction, deferring ambiguous and low-confidence samples to human experts. 

\begin{figure}[!ht]
\centering
\begin{minipage}{0.32\textwidth}
  \centering
  \includegraphics[width=\linewidth]{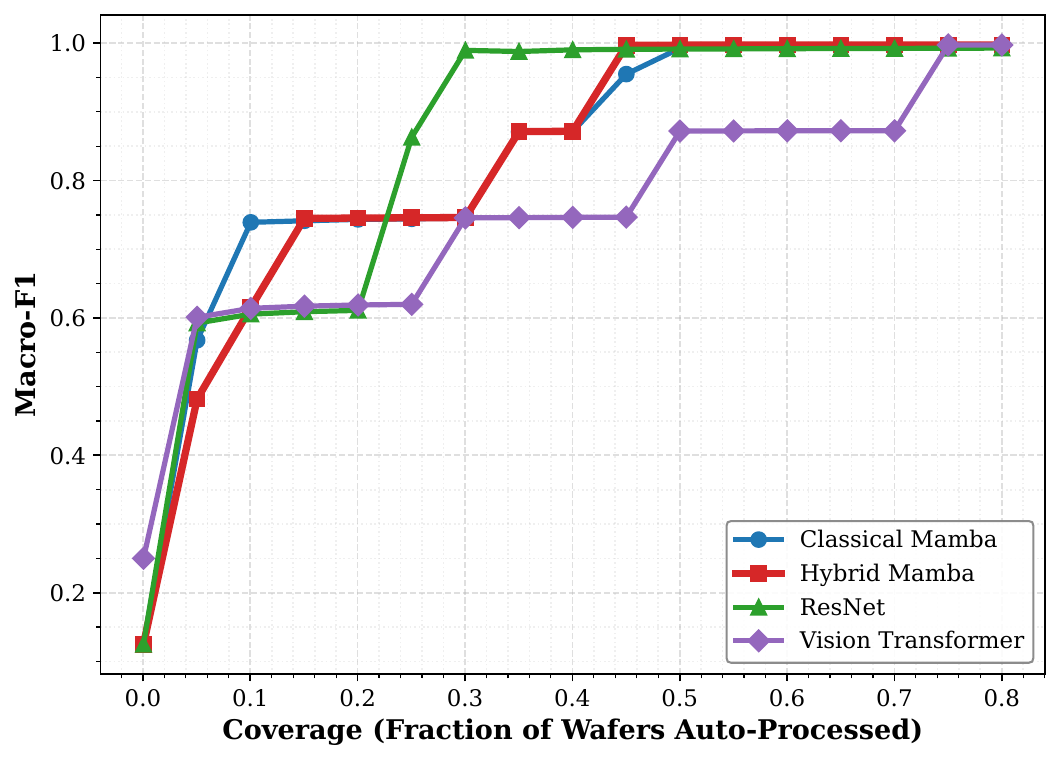}
  \caption{Selective Prediction Macro-F1 improvement.}
  \label{fig:sel_pred}
\end{minipage}\hfill
\begin{minipage}{0.32\textwidth}
  \centering
  \includegraphics[width=\linewidth]{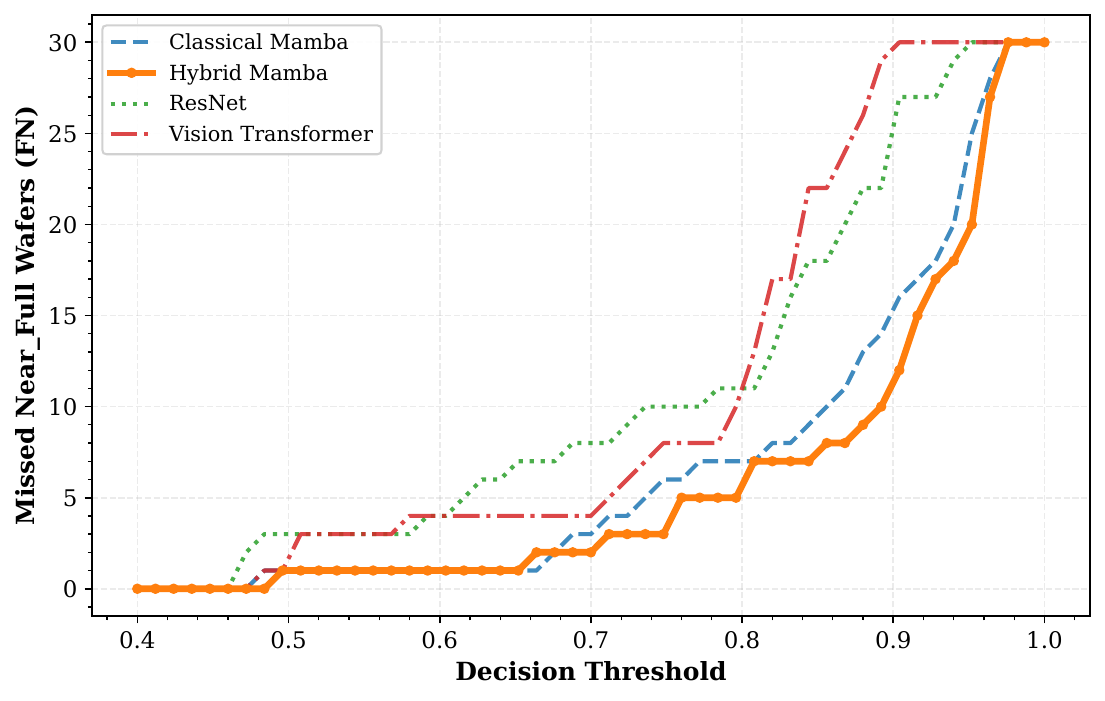}
  \caption{Catastrophic Miss Rate across decision thresholds.}
  \label{fig:miss_rate}
\end{minipage}\hfill
\begin{minipage}{0.32\textwidth}
  \centering
  \includegraphics[width=\linewidth]{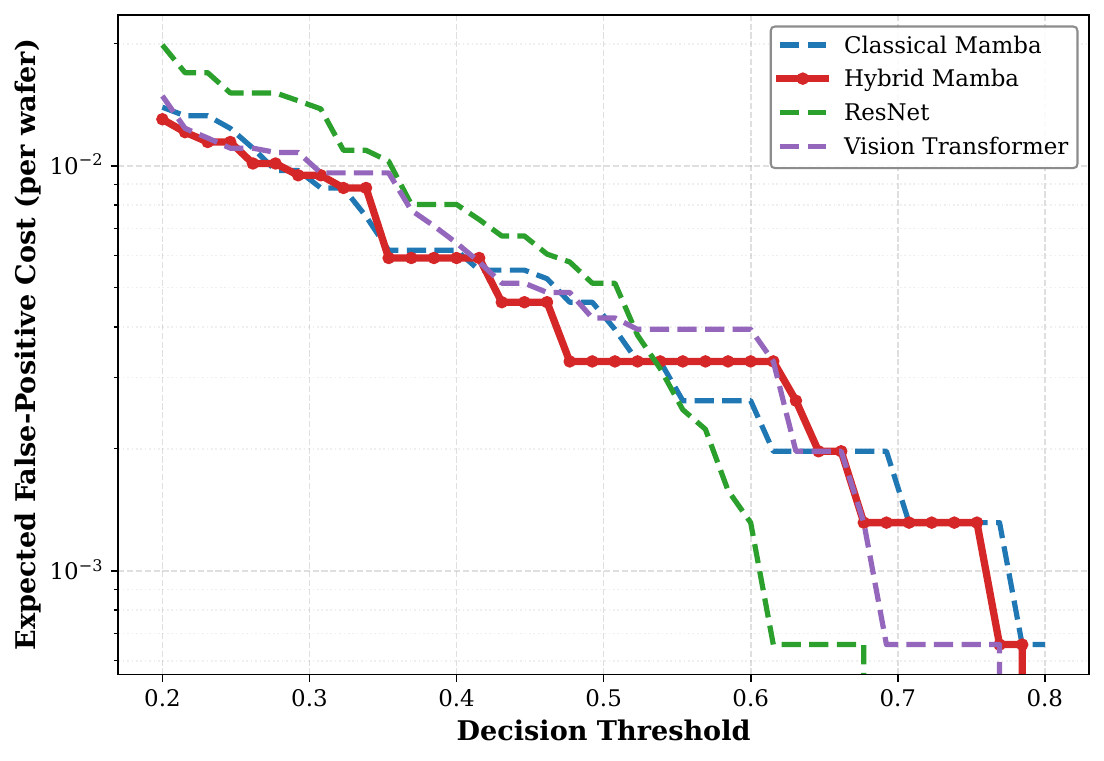}
  \caption{Expected False-Positive Cost reduction per wafer.}
  \label{fig:op_risk}
\end{minipage}
\end{figure}

As automated coverage decreases, the Hybrid Mamba reaches near-perfect Macro-F1 faster than classical baselines (Fig.~\ref{fig:sel_pred}) with a value of 0.9994. From the catastrophic miss rate (Fig.~\ref{fig:miss_rate}), we observe that the Hybrid model sustains zero missed `Near\_Full' defects at much higher confidence thresholds. This algorithmic calibration directly translates to financial value, minimizing the Expected False-Positive Cost per wafer (Fig.~\ref{fig:op_risk}) across all operating regimes.

Regarding practical deployment in high-throughput Automated Optical Inspection (AOI) systems, the architectural design explicitly prioritizes computational efficiency. By operating the 4-qubit Quantum Context Adapter strictly at the deepest semantic bottleneck, the quantum simulation processes a heavily compressed latent representation. This strategic placement ensures that the hybrid architecture avoids the prohibitive quadratic $\mathcal{O}(N^2)$ memory and computational bottlenecks inherent to standard Vision Transformers. 
Consequently, the model preserves the strictly linear $\mathcal{O}(N)$ scaling profile of the Mamba backbone, delivering the high-throughput, rapid inference speeds necessary for real-time anomaly detection on active fabrication lines.

\section{Conclusion}
To advance scalable representation learning in industrial data mining, we introduce Hybrid Quantum-MambaVision. Integrating linear-time State-Space Models with a 4-qubit Quantum Context Adapter, our architecture disentangles complex, multi-label wafer map patterns without the quadratic overhead of Vision Transformers. Crucially, the quantum bottleneck acts as a powerful algorithmic calibrator. By dynamically routing spatial features according to topological complexity, the model drastically reduces Maximum Calibration Error and prevents catastrophic misses on rare, highly imbalanced defect classes. Ultimately, this quantum-classical paradigm delivers a robust, high-throughput framework for real-time defect discovery and risk mitigation in semiconductor manufacturing.

\bibliographystyle{splncs04}
\bibliography{refrences}

\end{document}